\documentclass[sigconf]{acmart}

\usepackage{graphicx}
\usepackage{hyperref}
\usepackage{natbib}
\usepackage{subcaption}
\usepackage{xcolor}

\AtBeginDocument{%
  }

\copyrightyear{2026}
\acmYear{2026}
\setcopyright{cc}
\setcctype{by}
\acmConference[SIGSPATIAL '26]{The 34th ACM International Conference on Advances in Geographic Information Systems}{November 03--06, 2026}{Riverside, CA, USA}
\acmBooktitle{The 34th ACM International Conference on Advances in Geographic Information Systems (SIGSPATIAL '26), November 03--06, 2026, Riverside, CA, USA}
\acmDOI{10.1145/3841645.3844198}
\acmISBN{979-8-4007-2950-8/2026/11}

\begin{document}

\title[GeoNatureAgent Benchmark]{GeoNatureAgent Benchmark: Benchmarking LLM Agents for Environmental Geospatial Analysis Across Frontier and Open-Weight Foundation Models}

\author{Gabriel Diaz-Ireland}
\authornote{Corresponding author.}
\email{gdiazir1@jhu.edu}
\orcid{0009-0003-9049-4947}
\affiliation{%
  \institution{Universidad Cat\'{o}lica de \'{A}vila (UCAV)}
  \city{\'{A}vila}
  \country{Spain}
}
\affiliation{%
  \institution{Johns Hopkins University}
  \city{Baltimore, MD}
  \country{USA}
}

\author{Diego Prieto-Herr\'{a}ez}
\email{diego.prieto@ucavila.es}
\orcid{0000-0003-4636-2261}
\affiliation{%
  \institution{Universidad Cat\'{o}lica de \'{A}vila (UCAV)}
  \city{\'{A}vila}
  \country{Spain}}

\author{Mario Garc\'{\i}a Peces}
\email{mario.garcia.peces@gmail.com}
\affiliation{%
  \institution{Independent Researcher}
  \city{Madrid}
  \country{Spain}
}

\author{Javier Vel\'{a}zquez}
\email{javier.velazquez@ucavila.es}
\orcid{0000-0002-9188-3827}
\affiliation{%
  \institution{Universidad Cat\'{o}lica de \'{A}vila (UCAV)}
  \city{\'{A}vila}
  \country{Spain}}

\author{Devika Jain}
\email{kakkar@fas.harvard.edu}
\orcid{0000-0001-7800-929X}
\affiliation{%
  \institution{Center for Geographic Analysis, Harvard University}
  \city{Cambridge, MA}
  \country{USA}
}

\renewcommand{\shortauthors}{Diaz-Ireland et al.}

\begin{abstract}
	Environmental scientists spend disproportionate effort on data wrangling rather than analysis. New AI agents can be a helpful tool, but no benchmark exists to evaluate AI agents that automate environmental geospatial workflows through structured tool calling against real APIs.
	We introduce the GeoNatureAgent Benchmark, the first benchmark for environmental analysis agents that operate via structured tool calls to a production-style geospatial API. The benchmark comprises 93 tasks across 18 categories.
	Tasks are evaluated against an open, self-hostable geospatial API that serves three environmental indicators across Spain and Portugal via sixteen tools. We evaluate nine frontier and open-weight LLMs, reporting capability and per-case cost as orthogonal axes.
	Results manifest that (1)~Claude~Sonnet~4 achieves the highest capability at $60.8\%\pm0.8\%$, followed closely by DeepSeek~V3.2 at $56.3\%\pm3.1\%$, while no other model exceeds $51\%$; (2)~the cost-accuracy Pareto frontier is occupied mostly by open-weight models, with DeepSeek~V3.2 offering $93\%$ of Claude's capability at 11.6$\times$ lower cost; and (3)~structured tool calling against a real API provides a more discriminative measure of real-world agent capability, with mean accuracies $25$--$35$ percentage points below those reported on general-purpose GIS benchmarks.
\end{abstract}




\maketitle

\section{Introduction}
\label{sec:intro}

Environmental monitoring at scale requires multi-temporal geospatial analysis, including the detection of land cover change, the assessment of erosion risk, or the computation of vegetation indices; furthermore, it requires the generation of statistical summaries for regulatory compliance. Practitioners routinely spend the majority of their effort on data wrangling, discovering datasets, reprojecting coordinate systems, and debugging code against remote sensing APIs~\cite{zhang2023geogpt, li2023llmgeo}. This expertise barrier limits the adoption of geospatial methods by domain scientists who understand environmental systems but lack GIS programming skills.

Recent advances in large language models (LLMs) have led to the development of AI agent systems that translate natural language (NL) into executable geospatial operations. Architectures range from LLMs with tool pools~\citep{zhang2023geogpt, akinboyewa2024giscopilot} to fine-tuned specialists~\citep{zhang2024gtchain, zhang2025envgpt} to multi-agent systems achieving $85$--$97\%$ accuracy~\citep{luo2025geojsonagents, lee2025geollmsquad}.

However, a critical evaluation gap persists. Public benchmarks for LLM agents in geospatial settings remain scarce, and the few that exist target general GIS tasks (GeoBenchX~\citep{krechetova2025geobenchx}, ThinkGeo~\citep{shabbir2025thinkgeo}) rather than environmental science workflows. Environmental knowledge benchmarks (EnviroExam~\citep{huang2024enviroexam}) evaluate knowledge, not agent behavior. Code-generation benchmarks (UnivEARTH~\citep{kao2025univearth}) do not reflect the structured API interaction that production systems employ. At the foundation model level, GEO-Bench~\citep{lacoste2023geobench} and GEO-Bench-2~\citep{simumba2025geobench2} evaluate pixel-level vision models, not the agent-level tool orchestration that determines real-world usability.

This work fills this gap through three main contributions: (1)~GeoNatureAgent Benchmark is the first benchmark for environmental analysis agents using structured tool calling against a production-style geospatial API, comprising 93 tasks across 18 categories with a sixteen-tool agent interface, evaluated by eight mechanistic checks per case (no LLM-as-judge); (2)~a systematic comparison of nine LLMs spanning closed and open-weight models reports capability and per-case cost as orthogonal axes; and (3)~a cost-efficiency Pareto analysis reveals that the cost-accuracy Pareto frontier is occupied mostly by open-weight models.

\section{Related Work}
\label{sec:related}

Agent tool-use can be evaluated by GeoBenchX~\citep{krechetova2025geobenchx}, ThinkGeo~\citep{shabbir2025thinkgeo}, and UnivEARTH~\citep{kao2025univearth}. GeoBenchX is the closest comparable work to the GeoNatureAgent Benchmark, but uses low-level primitives on local files rather than structured tool calls against a production-style API.

GEO-Bench~\citep{lacoste2023geobench} and GEO-Bench-2~\citep{simumba2025geobench2} evaluate pixel-level model accuracy; GeoNatureAgent Benchmark evaluates agent-level tool orchestration. Both are complementary.

\section{GeoNatureAgent Benchmark}
\label{sec:benchmark}

\subsection{Benchmark Design}

GeoNatureAgent Benchmark comprises 93 tasks organized into 18 categories. The category structure is a capability taxonomy: each category isolates one competency required for environmental geospatial analysis---tool selection, spatial reasoning, multi-turn memory, error handling and recovery, cross-indicator synthesis, interpretation, and multilingual understanding, among others. Task counts are weighted by capability importance and coverage rather than balanced: tool selection receives the most tasks ($21$) because choosing the right operation from a realistic toolset is the pivotal agentic skill and the most discriminative across models (Section~\ref{sec:results}), whereas niche capabilities such as cross-country temporal context receive one or two.

Each task specifies a natural language query (optionally with multi-turn history), expected tool calls, must-contain/must-not-contain strings, maximum rounds, cost budget, and domain-expert ground truth. Error-handling tasks are deliberately unsolvable ---requesting analysis over non-existent municipalities, fabricating statistics from display-only layers, or citing indicators that do not exist --- testing the agent's ability to decline gracefully rather than hallucinate.

\paragraph{Worked example.} Case \texttt{V5\_21} (comparison): \emph{``Compare CO$_2$ geological storage suitability between Murcia and C\'{o}rdoba. Which province is better suited?''} The gold solution issues one \texttt{compare\_areas} call, returning Murcia $68.5\%$ vs.\ C\'{o}rdoba $68.4\%$; the expected answer reports both values and notes they are within tolerance.

The benchmark covers three environmental indicators across two countries: (1)~\textbf{CO$_2$ absorption suitability}~\citep{spain2025rd214} (Spain, categorical ---Not Eligible, Eligible with Conditions, Eligible---, served as a Cloud Optimized GeoTIFF (COG)), (2)~\textbf{Gully erosion probability}~\citep{borrelli2025gully} (Europe, continuous ---$0$--$100\%$---, served as a COG), and (3)~\textbf{BigEarthNet~V2 land cover}~\citep{clasen2024reben} (Portugal, categorical ---7-class land cover distribution---, served as pre-computed JSON statistics). Seven additional layers are available for visual display but not for statistical analysis, creating natural hallucination traps.

The benchmark's tools vary in geographic coverage: the CO$_2$ and gully-erosion indicators (raster COGs, Spain/Europe) are served by the full toolset, whereas the BigEarthNet land-cover indicator (pre-computed district statistics, Portugal) is served only by \texttt{analyze\_area}.

For direct comparison with general-purpose GIS benchmarks, the toolset integrates four canonical GIS operations from GeoBenchX~\citep{krechetova2025geobenchx}: proximity buffering, spatial-relationship selection, centroid extraction, and explicit task rejection for out-of-scope queries.

\subsection{Agent Architecture}

\begin{figure}[ht]
	\centering
	\includegraphics[trim={0.7cm 0.8cm 0.7cm 1.5cm}, clip, width=0.88\columnwidth]{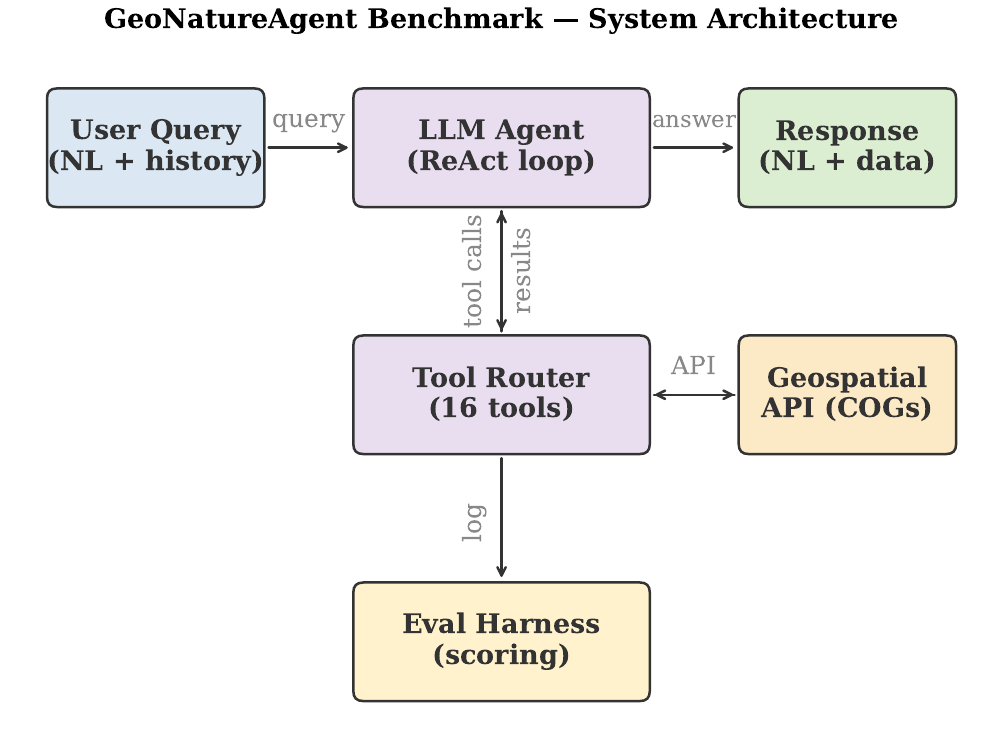}
	\caption{GeoNatureAgent Benchmark system architecture. NL denotes natural language.}
	\Description{}
	\label{fig:architecture}
\end{figure}

Figure~\ref{fig:architecture} shows the system architecture. The agent operates in a ReAct-style loop~\citep{yao2023react}: it receives a query, reasons about tools, executes calls, observes results, and generates a response. The agent has access to sixteen tools: the twelve principal operations plus four auxiliary tools for layer discovery and erosion time-series statistics. Tools are published via a Model Context Protocol (MCP) server exposing only tool primitives~\citep{mcp}.

\section{Experiments and Results}
\label{sec:results}

We evaluate nine LLMs on the GeoNatureAgent Benchmark v5 across Vertex~AI Model-as-a-Service, OpenRouter, and the Anthropic Messages API. To control for run-to-run variance, every model is evaluated under three random seeds $\{42, 1337, 2024\}$. Table~\ref{tab:leaderboard} summarizes the main results aggregated across seeds. Three findings stand out: (1)~the top of the leaderboard is bimodal, (2)~accuracy tracks scale strongly within our small open-weight sample, and (3)~closed-source frontier models do not categorically dominate open-weight peers. No model reaches $65\%$, confirming that environmental geospatial tool orchestration remains a significant challenge.

\begin{table}[ht]
	\caption{Capability leaderboard}
	\label{tab:leaderboard}
	\centering
	\begin{tabular}{rlrrr}
		\toprule
		\textbf{Model} & \textbf{Params} & \textbf{Accuracy} & \textbf{Cost} & \textbf{\$/case} \\
		\midrule
		Claude Sonnet 4 & --- & $60.8\% \pm 0.8\%$ & $\$11.84$ & $0.127$ \\
		DeepSeek V3.2  & 671B MoE & $56.3\% \pm 3.1\%$ & $\$1.05$  & $0.011$ \\
		GLM-5            & --- & $50.2\% \pm 2.2\%$ & $\$3.58$  & $0.038$ \\
		Gemini 2.5 Pro   & --- & $48.0\% \pm 3.3\%$ & $\$4.87$  & $0.052$ \\
		GPT-4o           & --- & $41.6\% \pm 2.7\%$ & $\$6.49$  & $0.070$ \\
		Qwen3-235B       & 235B MoE & $41.2\% \pm 4.3\%$ & $\$0.89$  & $0.010$ \\
		GPT-OSS-120B     & 120B & $34.1\% \pm 1.2\%$ & $\$8.27$  & $0.089$ \\
		Llama 4 Scout    & 109B MoE & $26.9\% \pm 2.1\%$ & $\$0.32$  & $0.003$ \\
		Gemma-3-27B      & 27B & $15.8\% \pm 1.6\%$ & $\$5.81$  & $0.062$ \\
		\bottomrule
	\end{tabular}
\end{table}

\begin{figure}[ht]
	\centering
	\includegraphics[trim={0.1cm 0.2cm 0.4cm 0.8cm}, clip, width=0.92\columnwidth]{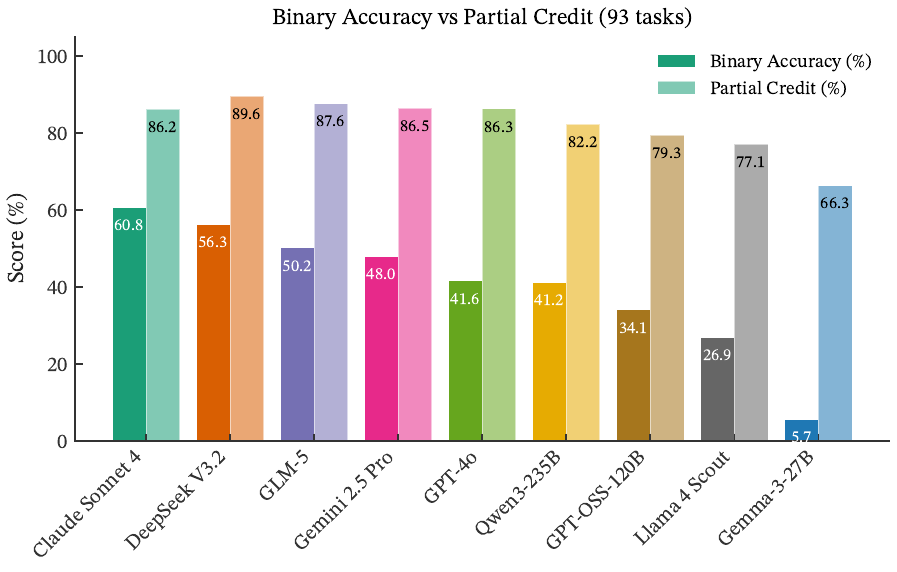}
	\caption{Binary accuracy vs partial-credit check score.}
	\Description{}
	\label{fig:binary_partial}
\end{figure}

Failing cases often demonstrate partial competence: correct tool selection but missing a keyword, or correct keywords but an extra unnecessary tool call. This gap between binary accuracy and partial-credit scores (see Figure~\ref{fig:binary_partial}) suggests many failures are ``near-misses'' addressable through prompt engineering.

\begin{figure}[ht]
	\centering
	\includegraphics[trim={0.1cm 0.1cm 1.8cm 1.9cm}, clip, width=0.92\columnwidth]{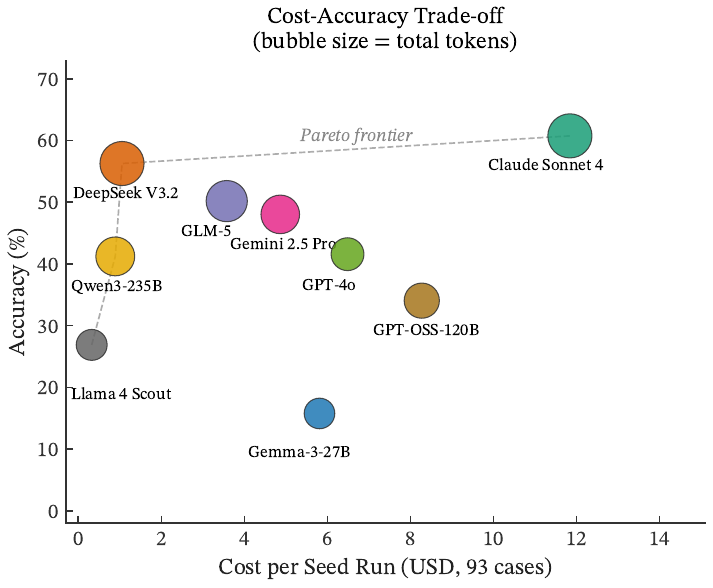}
	\caption{Cost-accuracy trade-off.}
	\Description{}
	\label{fig:cost_accuracy}
\end{figure}

Cost per case varies by more than an order of magnitude across the different models (see Table~\ref{tab:leaderboard}). The cost-accuracy Pareto frontier (Figure~\ref{fig:cost_accuracy}) comprises four models: \textbf{Llama~4~Scout} (cheapest), \textbf{Qwen3-235B}, \textbf{DeepSeek~V3.2}, and \textbf{Claude~Sonnet~4} (highest capability). Three of the Pareto-frontier models are open-weight; the remaining five models are all dominated. The DeepSeek~V3.2 entry is particularly notable: it delivers $93\%$ of Claude's capability at $11.6\times$ lower per-case cost, making it the strongest single-model recommendation for cost-sensitive deployments.

Extending the cost analysis to the environmental footprint, we estimate energy and carbon per query from token usage (blended $0.40$~Wh per $1{,}000$ tokens; global-average grid intensity $400$~gCO$_2$/kWh; an order-of-magnitude estimate, not a measurement): the per-case footprint broadly rises with capability, from $3.1$~gCO$_2$ (Gemma-3-27B) to $6.4$~gCO$_2$ (Claude~Sonnet~4).

\subsection{Failure Analysis}
\label{sec:heatmap}

\begin{figure}[ht]
	\centering
	\includegraphics[trim={0.1cm 0.2cm 0.1cm 0.6cm}, clip, width=0.92\columnwidth]{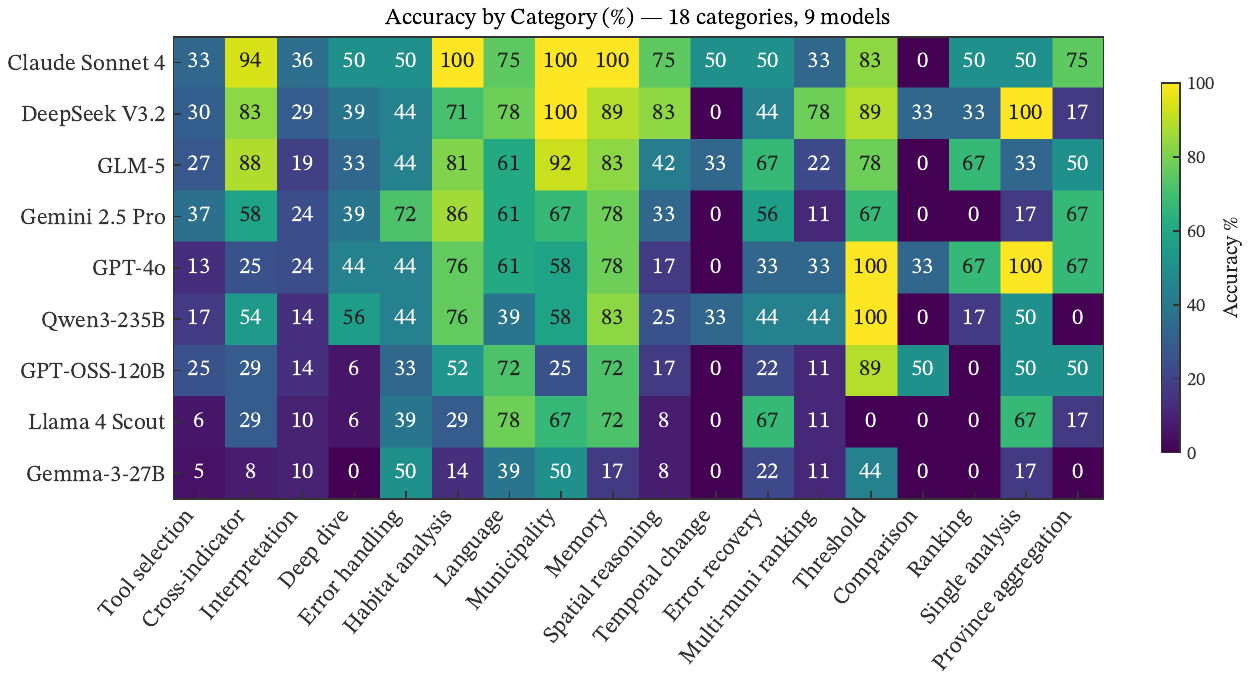}
	\caption{Mean accuracy by category across the nine models.}
	\Description{}
	\label{fig:heatmap}
\end{figure}

Accuracy varies substantially across different categories, revealing that models handle certain tasks more effectively than others (see Figure~\ref{fig:heatmap}). Tasks requiring contextual retention or multilingual parsing prove relatively accessible, such as multi-turn memory (cohort mean $75\%$), habitat analysis ($65\%$), and language understanding ($63\%$). In contrast, models fail in tasks demanding multi-step numeric reasoning and precise tool orchestration, such as ranking ($26\%$), comparison ($13\%$), and tool selection ($22\%$).  Furthermore, models differ in consistency: while some exhibit severe competency gaps, Claude~Sonnet~4 and DeepSeek~V3.2 maintain a more balanced profile.

Across all $1{,}429$ failed model--seed--case runs, failures concentrate in tool selection rather than data handling. Over two-thirds of all errors ($68.0\%$) stem from missing tool calls, while only $9.7\%$ are caused by extracting wrong data, indicating the bottleneck lies in tool routing rather than data interpretation.

\section{Discussion}
\label{sec:discussion}

GeoNatureAgent Benchmark's $60.8\%$ best-model accuracy is notably lower than GeoBenchX and GeoJSON~Agents ($85$--$97\%$).  Four factors explain the accuracy gap: (1)~real API interaction vs.\ local file operations; (2)~environmental domain complexity; (3)~multi-turn conversation history; and (4)~strict binary evaluation. Our results align with harder benchmarks: MapEval ($<67\%$)~\citep{dihan2025mapeval}, SpatiaLab ($55\%$)~\citep{wasi2026spatialab}, and UnivEARTH ($33\%$)~\citep{kao2025univearth}.

The structured approach also constrains the output space, eliminating the $58\%$ code execution failure rate that UnivEARTH reports~\citep{kao2025univearth} ---further evidence that production environmental systems should prefer structured APIs over free-form code generation.

\section{Conclusion}
\label{sec:conclusion}

GeoNatureAgent Benchmark is developed for evaluating AI agents on environmental geospatial analysis through structured tool calling against an open, self-hostable API. The benchmark spans three indicators across Spain and Portugal, demonstrating domain and geographic extensibility. Evaluating nine LLMs on $93$ tasks across 18 categories with sixteen tools, it can be concluded:

\begin{enumerate}
    \item \textbf{Environmental geospatial tool orchestration remains open.} The capability leader (Claude~Sonnet~4) achieves $60.8\% \pm 0.8\%$, well below the $85$--$97\%$ reported on general GIS benchmarks.
    \item \textbf{Open-weight models occupy the cost-accuracy Pareto frontier.} Llama~4~Scout, Qwen3-235B, and DeepSeek~V3.2.
    \item \textbf{The capability-cost trade-off is steep.} DeepSeek~V3.2 delivers $93\%$ of Claude's capability at $11.6\times$ lower per-case cost.
    \item \textbf{Systematic failures persist.} Tool-selection variance remains the dominant inter-model spread, and parameter scale alone does not predict capability.
\end{enumerate}

\begin{acks}
We thank Darwin Geospatial for infrastructure support for this research and for open-sourcing the code and experiments. The Portuguese land-cover statistics derive from BigEarthNet~V2, which contains modified Copernicus Sentinel data (2017--2018); we acknowledge the European Space Agency and the European Commission Copernicus programme. We also thank Guillermo Perez for his contributions during his internship at Darwin Geospatial.
\end{acks}

\bibliographystyle{ACM-Reference-Format}
\bibliography{references}

\appendix

\section{Online Resources}
\label{sec:Data_Availability}

Code, evaluation harness, and self-hostable API: \url{https://github.com/gabrielireland/GeoNatureAgent_Benchmark}. Benchmark dataset and evaluation software: Zenodo DOIs 10.5281/zenodo.20450995~\citep{geonatureagent_dataset} and 10.5281/zenodo.20450997~\citep{geonatureagent_software}.

\end{document}